\documentclass[10pt,twocolumn,letterpaper]{article}

\usepackage{cvpr}
\usepackage{times}
\usepackage{epsfig}
\usepackage{graphicx}
\usepackage{amsmath}
\usepackage{amssymb}
\usepackage{booktabs}
\usepackage{multirow}
\usepackage{color}
\usepackage{pifont}
\usepackage[dvipsnames]{xcolor}
\usepackage{enumitem}
\usepackage{authblk}

\definecolor{orange}{RGB}{255,127,0} 
\definecolor{darkgreen}{RGB}{0, 100, 0}

\usepackage[pagebackref=true,breaklinks=true,letterpaper=true,colorlinks,bookmarks=false]{hyperref}

\cvprfinalcopy 

\def\cvprPaperID{6204} 

\ifcvprfinal\fi
\begin{document}

\title{Action Modifiers: Learning from Adverbs in Instructional Videos}
\renewcommand*{\Authsep}{\qquad}
\renewcommand*{\Authand}{\qquad}
\renewcommand*{\Authands}{\qquad}
\makeatletter
\renewcommand\AB@affilsepx{\qquad \protect\Affilfont}
\makeatother
\author[1]{Hazel Doughty}
\affil[1]{University of Bristol}

\author[2]{Ivan Laptev}
\affil[2]{Inria, \'{E}cole Normale Sup\'{e}rieure}
\author[1,3]{Walterio Mayol-Cuevas}
\affil[3]{Amazon}
\author[1]{Dima Damen}


\maketitle
\thispagestyle{empty}

\begin{abstract}
\vspace{-0.6em}
We present a method to learn a representation for adverbs 
from instructional videos using weak supervision from the accompanying narrations. 
Key to our method is the fact that the visual representation of the adverb is highly dependant on the action to which it applies, although the same adverb will modify multiple actions in a similar way.
For instance, while `spread quickly' and `mix quickly' will look dissimilar, we can learn a common representation that allows us to recognize both, among other actions.

We formulate this as an embedding problem, and use scaled dot-product attention to learn from weakly-supervised video narrations. We jointly learn adverbs as invertible transformations 
operating on the embedding space, so as to add or remove the effect of the adverb. 
As there is no prior work on weakly supervised learning of adverbs, we gather paired action-adverb annotations from a subset of the HowTo100M dataset for 6 adverbs: \textit{quickly/slowly, finely/coarsely, and partially/completely}. 
Our method outperforms all baselines for video-to-adverb retrieval with a performance of 0.719 mAP. We also demonstrate our model's ability to attend to the relevant video parts in order to determine the adverb for a given action.
\end{abstract}

\vspace{-1em}
\section{Introduction}
Instructional videos are a popular type of media watched by millions of people around the world to learn new skills. 
Several previous works aimed to learn the key steps necessary to complete the task from these videos~\cite{alayrac2016unsupervised, malmaud2015s, sener2015unsupervised, zhukov2019cross}. However, identifying the steps, or their order, is not all one needs to perform the task well; 
some steps need to be performed in a certain way to achieve the desired outcome. Take for example the task of making a meringue. An expert would assure you it is critical to add the sugar \textit{gradually} and avoid over-beating by folding the mixture \textit{gently}. 

\begin{figure}
    \centering
    \includegraphics[width=0.95\linewidth]{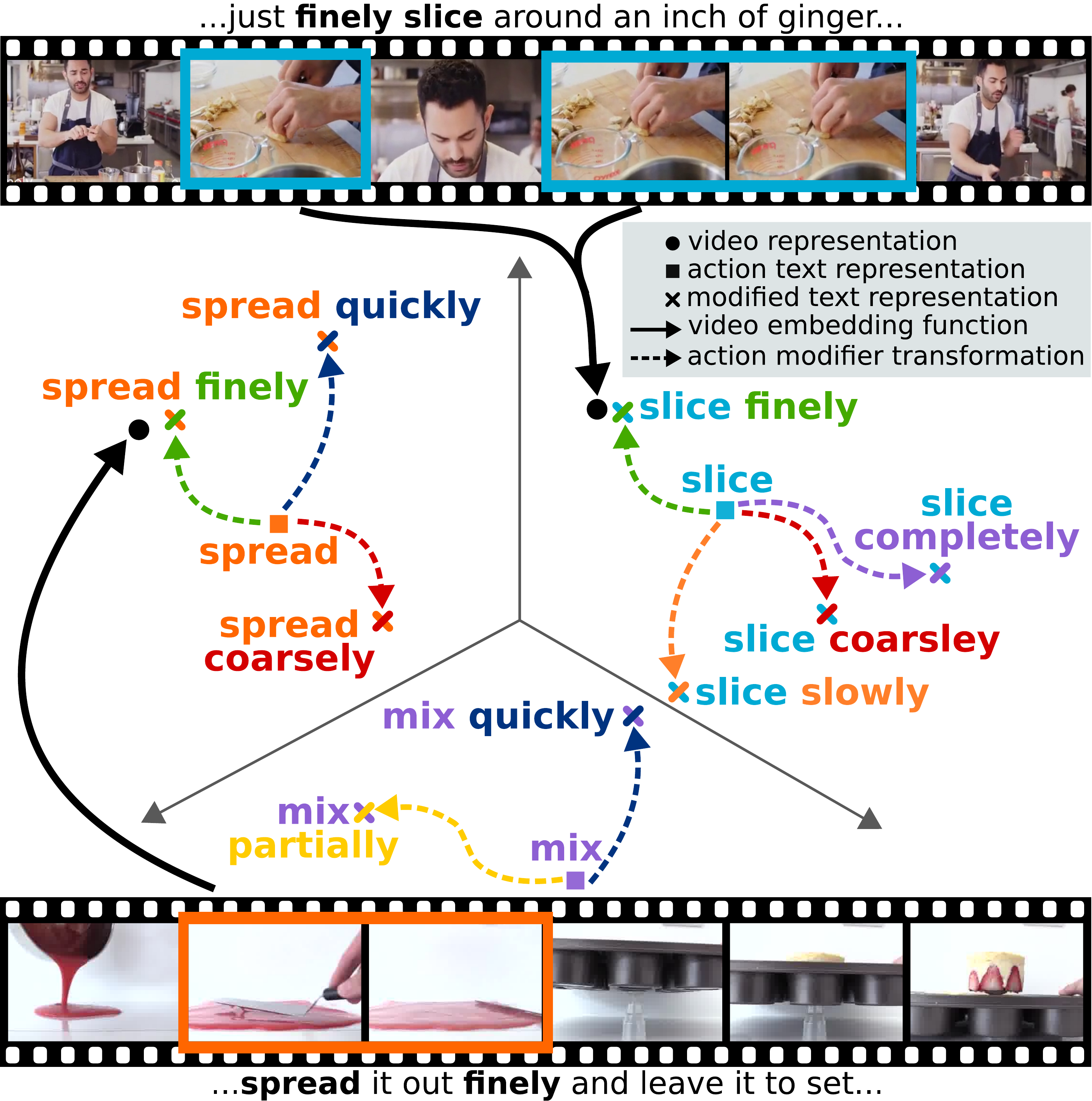}
    \caption{
    We learn a joint video-text embedding space from instructional videos and accompanying action-adverb pairs in the narration.
    Within this space, we learn adverbs as action modifiers --- that is transformations which modify the action's embedding. 
    }
    \vspace{-.3cm}
    \label{fig:concept}
\end{figure}

This is related to recent efforts on assessing the performance of daily tasks~\cite{doughty2018s, doughty2019pros, li2019manipulation}, however, these works do not assess individual actions or identify whether they have been performed as recommended by, say, a recipe. 
As in the example before, steps with such caveats 
are often indicated by adverbs describing how actions should be performed.
These adverbs (\eg~\textit{quickly, gently, ...}) generalize to different actions and modify the manner of an action. We thus learn these as \textbf{action modifiers} (Fig. ~\ref{fig:concept}).

To learn action modifiers for a variety of tasks and actions, we utilize the online resource of instructional videos with accompanying narrations. 
However, this form of supervision is weak and noisy.
Not only are the narrations just roughly aligned with the actions in the video, but often the narrated actions may not be captured in the video altogether. 
For example, a YouTube instructional video might be narrated as 
``pour in the cream quickly'' but the visuals only show the cream already added. In this case the video would not be useful to learn the adverb `quickly'.

As the main contribution of this paper, we propose the first method for weakly supervised learning from adverbs, in which we embed relevant video segments in a latent space and learn adverbs as transformations in this space. We collect action-adverb labels from narrations of a subset of tasks in the HowTo100M dataset~\cite{miech2019howto100m}. 
The method is evaluated for video-to-adverb retrieval,
as well as adverb-to-video retrieval and shows significant improvements over baselines. 
Additionally, we present a comprehensive ablation study demonstrating that jointly learning a good action embedding is key to learning action modifiers.


\section{Related Work}
\label{sec:rel}
We review works which learn from instructional videos, followed by works using parts-of-speech in video. We then review the related task of object attributes in images and methods which learn embeddings under weak supervision. 

\medskip
\noindent \textbf{Instructional Videos.} 
Movies accompanied by subtitles and scripts have been used for learning from video~\cite{duchenne2009automatic, everingham2006hello, laptev2008learning,tapaswi2015Book2Movie}. 
However, movies typically focus on talking heads with few object interactions.
More recently, instructional videos are a popular source of datasets~\cite{alayrac2016unsupervised, miech2019howto100m, sener2019zero, zhou2018towards} 
with hundreds of online videos of the same task. 
Narrations are used to learn steps of complex tasks~\cite{alayrac2016unsupervised, huang2016connectionist, malmaud2015s, richard2018action, sener2015unsupervised, zhukov2019cross}, and more recently for video retrieval~\cite{miech2019howto100m}, visual grounding~\cite{huang2018finding, huang2017unsupervised}, action segmentation~\cite{zhou2018towards} and learning actions through object state changes~\cite{alayrac2017joint, fathi2013modeling}. 

In this work, we offer a novel insight into how these instructional videos can be used beyond step identification. Our work utilizes videos from the recently released HowTo100M dataset~\cite{miech2019howto100m}, learning adverbs and their relevance to critical steps in these tasks. 

\medskip
\noindent \textbf{Learning from Parts-of-Speech in Video.} 
Several problems are at the intersection between language and video: captioning~\cite{krishna2017dense, pan2017video, yao2015describing, zeng2016generation}, retrieval~\cite{dong2019dual, hendricks2017localizing, jain2015objects2action, mettes2017spatial, miech2019howto100m, wray2019fine, xu2015jointly} and visual question answering~\cite{gao2018motion, yu2017multi, yu2017end, zhu2017uncovering}. The majority of these works use LSTMs or GRUs to combine words into sentence-level features. 
While some works use learned pooling~\cite{miech2017learnable} or attention~\cite{yao2015describing, yu2017multi, yu2017end}, they do not use knowledge of the words' parts-of-speech (PoS). 

A few recent works differentiate words by their PoS tags. Xu et al.~\cite{xu2015jointly} learn a joint video-text embedding space after detecting (subject, verb, object) triplets in the input caption. Wray et al.~\cite{wray2019fine} perform fine-grained action retrieval by learning a separate embedding for each PoS before combining these embeddings. 
Both works focus on verb and noun PoS, as they target action recognition. Alayrac et al.~\cite{alayrac2016unsupervised} also use verb-noun pairs; the authors use direct object relations to learn unsupervised clusterings of key steps in instructional videos. 

While some adverbs are contained in video captioning  datasets~\cite{krishna2017dense, zeng2016generation}, no prior captioning work models or recognizes these adverbs. 
The only prior work to utilize adverbs is that of Pang et al.~\cite{pang2018human} where many adverbs in the ADHA dataset model moods and facial expressions (\eg `happily', `proudly'). The work uses full supervision including action bounding boxes. 
Instead, in this work we target adverbs that represent the manner in which an action is performed, using only weak supervision from narrations.  

\medskip
\noindent \textbf{Object Attributes in Images.} 
Adverbs of actions are analogous with adjectives of objects.
 Learning adjectives for nouns 
 has been investigated in the context of recognizing object-attribute pairs~\cite{borth2013large, chen2014inferring, isola2015discovering, misra2017red, nagarajan2018attributes, nan2019recognizing, wang2013unified, wang2010discriminative} from images. 
 Both~\cite{chen2014inferring,misra2017red} tackle the problem of contextuality of attributes, where the appearance of an attribute can vastly differ depending on the object it applies to. Chen and Grauman~\cite{chen2014inferring} formulate this as transfer learning to recognize unseen object-attribute compositions. Misra et al.~\cite{misra2017red} learn how to compose separate object and attributes classifiers for novel combinations.
 Instead of using classifiers to recognize attributes, Nagarajan and Grauman
 ~\cite{nagarajan2018attributes} model attributes as a transformation of an object's embedding. Our work is inspired by this approach. 

 While some works learn attributes for actions~\cite{liu2011recognizing, rosenfeld2018action, zellers2017zero}, these detect combinations of specific attributes (\eg~`outdoor', `uses toothbrush') to perform zero shot recognition and do not consider adverbs as attributes.

\medskip
\noindent \textbf{Weakly Supervised Embedding.} 
Learned embeddings are commonly used for retrieval tasks, however few works have attempted to learn embeddings under weak supervision~\cite{arandjelovic2016netvlad,mithun2019weakly,tan2019wman, xu2019weakly}. 
In~\cite{arandjelovic2016netvlad}, weak supervision is overcome using
a triplet loss that only optimizes distances to the definite negatives and identifies the best matching positive. 
Two works~\cite{mithun2019weakly,tan2019wman}
perform video moment retrieval from text queries without temporal bounds in training. Similar to our approach, both use a form of text-guided attention to find the relevant portion of the video, however these use the full sentence. In our work, we simultaneously embed the relevant portion of the video while learning how adverbs modify actions. We detail our method next.

\begin{figure*}[t]
    \centering
    \includegraphics[width=0.97\linewidth]{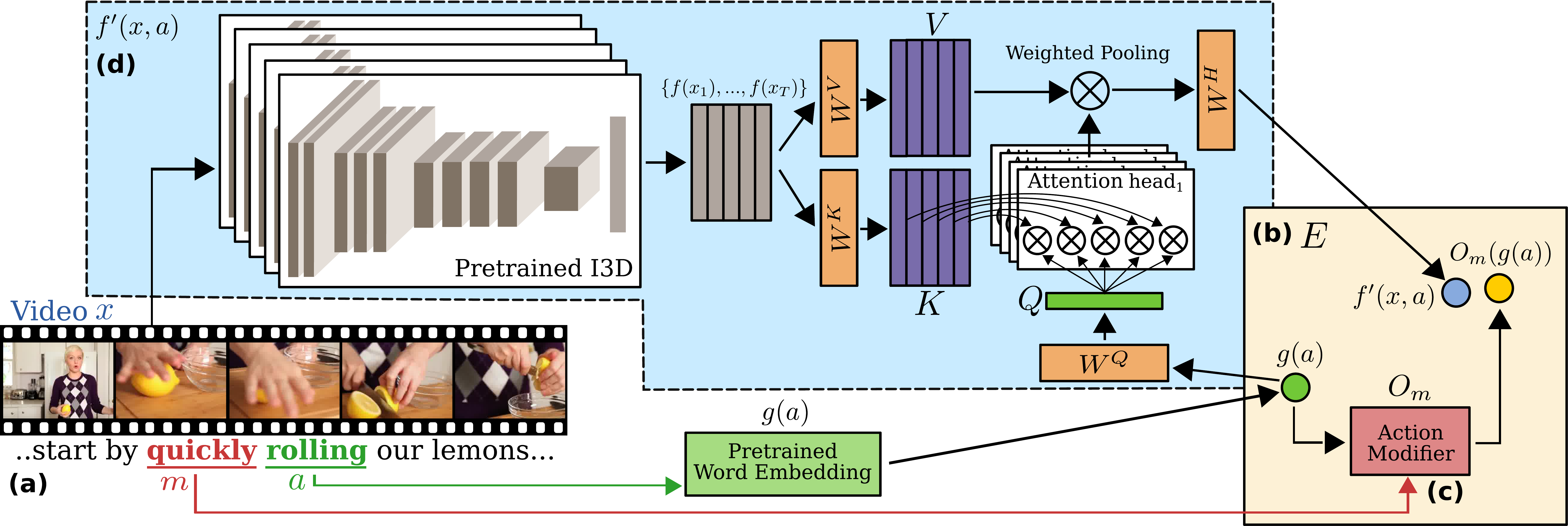}
    \caption{
    (a) Our input is a video $x$ with the weak label $(a,m)$ for the action and adverb respectively. (b) We aim to learn a joint video-text embedding space for adverb and video retrieval where the embedded video (blue) and action-adverb text representation (yellow) are close. (c) We learn adverbs as action modifiers which are transformations in the embedding space. (d) We embed $f'(x,a)$, a visual representation of the relevant video parts using multi-head scaled dot-product attention where the query is a projection of the action embedding $g(a)$. 
    }
    \vspace*{-6pt}
    \label{fig:method}
\end{figure*}

\section{Learning Action Modifiers}
\label{sec:method}


The inputs to our model are action-adverb narrations and the accompanying instructional videos. Fig.~\ref{fig:method}(a) shows a sample instructional video, narrated with ``...start by \textcolor{BrickRed}{quickly} \textcolor{ForestGreen}{rolling} our lemons...'', from which we identify the action \textcolor{ForestGreen}{roll} and the adverb \textcolor{BrickRed}{quickly} (see Sec.~\ref{sec:dataset} for NLP details). After training, our model is able to assess whether videos in the test set, of the same or different action, have been achieved \textcolor{BrickRed}{quickly}, among other learned adverbs.

We present an overview of our method in Fig.~\ref{fig:method}. We learn a joint video-text embedding shown in Fig.~\ref{fig:method}(b), where the 
relevant video parts are embedded~(blue dot) close to the text representation of the adverb-modified action `roll quickly' (yellow dot). 
We review how joint video-text embeddings are typically learned in Sec.~\ref{sec:action_emb}.
This section also introduces the notations for the rest of the paper.

Two prime challenges exist in learning the embedding for our problem, \ie learning from adverbs in instructional videos. The first is disentangling the representation of the action from the adverb, allowing us to learn how the same adverb applies across different actions. We propose to learn adverbs as action modifiers, one per adverb, as in Fig.~\ref{fig:method}(c).
In Sec.~\ref{sec:action_mods} we introduce these action modifiers, which we represent as transformations in the embedding space.

The second challenge is learning the visual representation from the relevant parts of the video in a weakly-supervised manner, \ie without annotations of temporal bounds.
In Sec.~\ref{sec:webly}, we propose a weakly-supervised embedding function that utilizes multi-head scaled dot-product attention.
This uses the text embedding of the action as the query to attend to relevant video parts, as shown in Fig.~\ref{fig:method}(d).


\subsection{Learning an Action Embedding}
\label{sec:action_emb}
Our base model is a joint video-text embedding, as in~\cite{miech2017learnable, wray2019fine, xu2015jointly}.
Specifically, given a set of video clips $x \in X$ 
with corresponding action labels $a \in A$, our goal is to obtain two embedding functions, one visual and one textual, $f: X \rightarrow E$ and $g: A \rightarrow E$ such that $f(x)$ and $g(a)$ are close in the embedding space $E$ and $f(x)$ is distant from other action embeddings $g(a')$. 
These functions $f$ and $g$ can be optimized with a standard cross-modal triplet loss:
\begin{align}
    \mathcal{L}_{triplet} = \mathrm{max}(0, &d(f(x), g(a)) \nonumber\\ - &d(f(x), g(a')) + \beta) \ s.t. \ a' \neq a
\end{align}
where $a'$ is an action different to $a$, $d$ is the Euclidean distance and $\beta$ is the margin, set to $1$ in all experiments. 
We use $g(a)$ as the GloVe~\cite{pennington2014glove} embedding of the action's verb, and explain how we replace $f(x)$ by $f'(x,a)$ in Sec.~\ref{sec:webly}.  

\subsection{Modeling Adverbs as Action Modifiers}
\label{sec:action_mods}
While actions exist without adverbs, adverbs are by definition tied to the action, and only gain visual representation when attached to one.
Although adverbs have a similar effect on different actions, the visual representation is highly dependent on the action.
Therefore, we follow prior work from~\cite{nagarajan2018attributes} on object-attribute pairs and model adverbs as learned transformations 
in the video-text embedding space $E$~(Sec.~\ref{sec:action_emb}). As these transformations modify the embedding of the action, we call them \textbf{action modifiers}. 
We learn an action modifier $O_m$ for each adverb $m \in M$, such that
\begin{equation}
\label{eq:actionmodifier}
    O_m(z) = W_m z 
\end{equation}
where $z$ is any point in the embedding space $E$ and ${O_m:E \rightarrow E}$ is a learned linear transform by a weight matrix $W_m$. In Sec.~\ref{sec:results}, we test other geometric transformations: fixed translation, learned translation and nonlinear transformation.
Each transformation $O_m$ can be applied to 
any text representation $O_m(g(a))$ or video representation $O_m(f(x))$ in $E$ to add the effect of the adverb $m$.

A video $x \in X$, labeled with action-adverb pair $(a, m)$, contains a visual representation of the adverb-modified action. We thus aim to embed $f(x)$ close to $O_m(g(a))$. This is equivalent to embedding the inverse of the transformation $O_m^{-1}(f(x))$ near the action~$g(a)$.
We thus jointly learn our embedding, with the action modifiers $O_m$, using the sum of two triplet losses. The first focuses on the action:
\begin{align}
    \mathcal{L}_{act} = \mathrm{max}(0, &d(f(x), O_m(g(a))) \nonumber\\ - &d(f(x), O_m(g(a'))) + \beta) \ s.t. \ a' \neq a
    \label{eq:lossaction}
\end{align}
where $a'$ is a different action and $d$ and $\beta$ are the distance function and margin as in Sec.~\ref{sec:action_emb}. Similarly, we have a triplet loss that focuses on the adverb, such that:
\begin{align}
    \mathcal{L}_{adv} = \mathrm{max}(0, &d(f(x), O_m(g(a))) \nonumber\\ - &d(f(x), O_{\overline{m}}(g(a))) + \beta)  
    \label{eq:lossadverb}
\end{align}
where $\overline{m}$ is the antonym of the labeled adverb $m$ (\eg when $m=$ `quickly', the antonym $\overline{m}=$ `slowly'). 
We restrict the negative in $\mathcal{L}_{adv}$ to only the antonym to deal with adverbs not being mutually exclusive. 
For instance, a video labeled `slice quickly' does not preclude the slicing being also done `finely'. However, it surely has not been done `slowly'. We demonstrate the effect of this choice in Sec.~\ref{sec:results}.

\subsection{Weakly Supervised Embedding}
\label{sec:webly}
All prior works that learn attributes of objects from images~\cite{chen2014inferring, isola2015discovering, misra2017red, nagarajan2018attributes,  nan2019recognizing} utilize fully annotated datasets, where the object 
the attributes relate to is the only object of interest in the image.
In contrast,
we aim to learn action modifiers from video in a weakly supervised manner. Our input is untrimmed videos containing multiple consecutive actions. 
To learn adverbs, we need the visual representation to be only from 
the video parts relevant to the action
(\eg `roll' in our Fig.~\ref{fig:method} example).
We propose using scaled dot-product attention~\cite{vaswani2017attention}, where the embedded action of interest acts as a query to identify relevant video parts.

For each video $x$, we use a temporal window of size~$T$, centered around the timestamp of the narrated action-adverb pair, containing 
video segments $\{x_1, x_2, ..., x_T\}$.
We start from the visual representation of all segments ${f(x) = \{f(x_1),...,f(x_T)\}}$, where $f(\cdot)$ is an I3D network.
From this, we wish to learn an embedding of the visual features relevant to the action $a$, which we call $f'(x,a)$.
Inspired by~\cite{vaswani2017attention}, we project $f(x)$ into keys $K$ and values $V$:
\begin{equation}
    K = W^K f(x); \qquad V = W^V f(x)
\end{equation}
We then set the query $Q = W^Q g(a)$ to be the projection of the action embedding, to weight video segments by their relevance to that action.
The attention weights are obtained from the dot product of the keys $K$ and the action query $Q$. These then pool the values $V$. Specifically:
\begin{equation}
    \label{eq:att}
    H(x,a) = \sigma\left( \frac{(W^Q g(a))^{\top}W^K f(x)}{\sqrt{T}}\right) W^V f(x)
\end{equation}
where $H(x,a)$ is a single attention head and $\sigma$ is the softmax function. We train multiple attention heads such that,
\begin{equation}
    f'(x,a) = W^H [H_1(x,a), ..., H_h (x,a)]
\end{equation}
where $W^H$ projects the concatenation of the multiple attention heads $H_i (x,a)$ into the embedding space. We learn $h$ attention head weights: $W_i^Q, W_i^K, W_i^V$ as well as $W^H$ parameters for our weakly-supervised embedding.

It is important to highlight that these weights are jointly trained with the embedding space $E$, so that $f'(x,a)$ is used instead of $f(x)$ in Equations~\ref{eq:lossaction} and~\ref{eq:lossadverb}. We opted to explain our embedding space before detailing how it can be learned in a weakly-supervised manner, to simplify the explanation.
 
\subsection{Weakly Supervised Inference}

Once trained, our model can be used to evaluate cross-modal retrieval of videos and adverbs. For video-to-adverb retrieval, we consider a video query $x$ and the narrated action $a$, and we wish to estimate the adverb $m$. For example, we have a video and wish to find the manner in which the action `slice' was performed. We use the learned function  $f'(x,a)$ to embed the relevant visual representation for action $a$ in $E$. We then rank adverbs by the distance from this embedding to all modified actions $\forall m: O_m(g(a))$.

For adverb-to-video retrieval, we consider an action-adverb pair $(a,m)$ as a query, embed $O_m(g(a))$, \eg `slice finely', and calculate the distance from this text representation to all relevant video segments $\forall x: f'(x,a)$. 
For both cases, this allows us to use $a$ to query to the weakly supervised embedding, so as to attend to the relevant video parts.

\begin{figure*}[t]
    \centering
    \includegraphics[width=\linewidth]{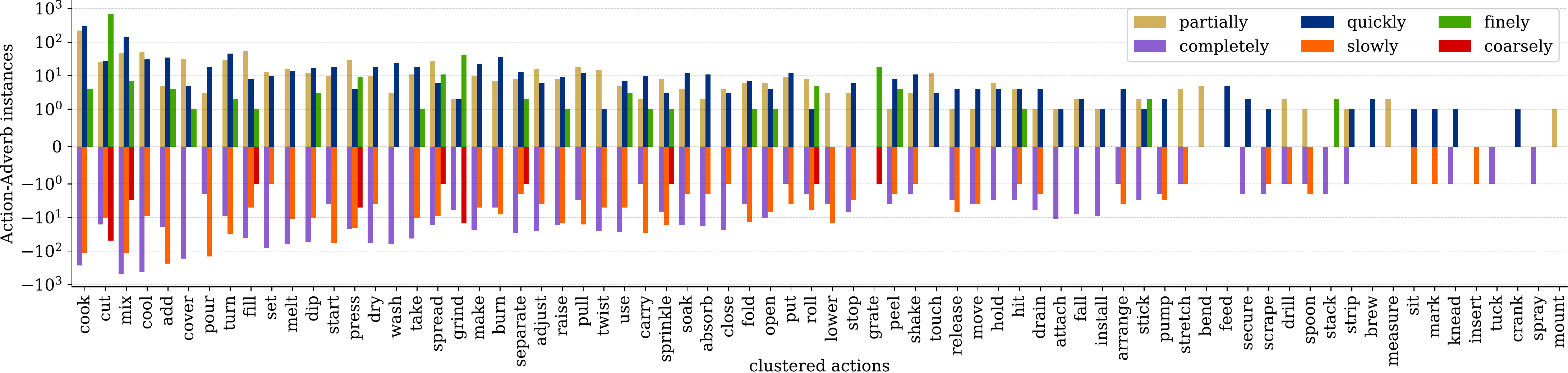}
    \caption{
    Log-scaled y-axis shows instances of each adverb plotted per action. We display adverbs against their paired antonym (+/- axis). 
    }
    \label{fig:action_dist}
    \vspace{-0.8em}
\end{figure*}

\begin{figure}
    \centering
    \includegraphics[width=\linewidth]{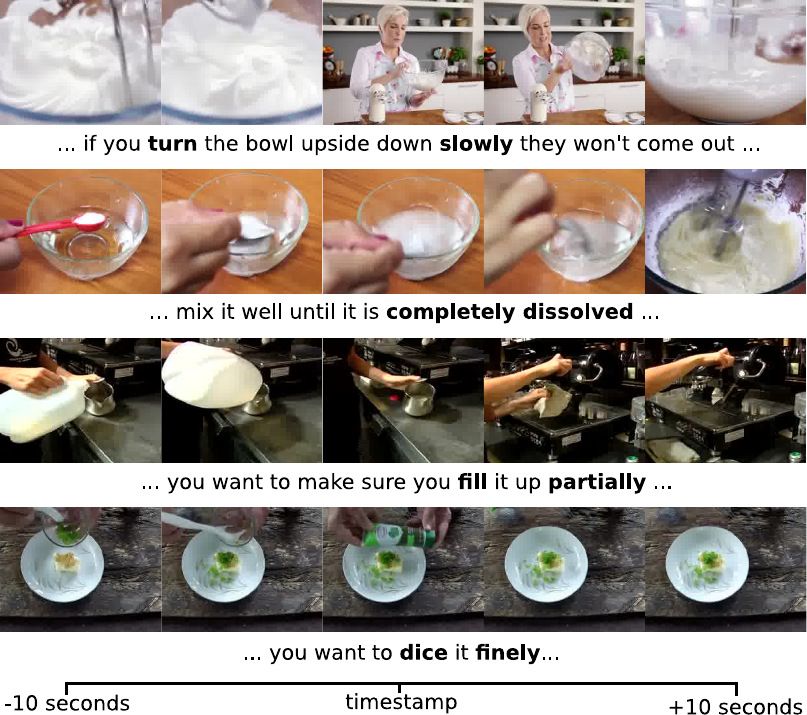}
    \caption{Example videos and narrations, highlighting the action and adverb discovered with our NLP pipeline. In some cases the weak timestamp is a good localization of the action (top), however in others the action is long (second), the timestamp is a poor match~(third), or the action is not captured in the video (bottom).}
    \vspace{-0.5em}
    \label{fig:dataset}
    
\end{figure}

\section{Dataset}
\label{sec:dataset}
HowTo100M~\cite{miech2019howto100m} is a large scale dataset of instructional videos collected from YouTube. Each video has a corresponding narration from manually-entered subtitles or Automatic Speech Recognition (ASR). No ground-truth is available in terms of correct actions or temporal extents. 

To test cross-task generalization, we use the same 83 tasks previously used in~\cite{zhukov2019cross}. These come from cooking, DIY and car maintenance, and are divided into 65 tasks for training and a disjoint set of 18 tasks for testing.
However, in~\cite{zhukov2019cross}, only 30 videos per task were used in training. Instead, we use all videos available for these 65 training tasks, where each task consists of 100-500 videos. 
In total, we have 24,558 videos in training and 1,280 videos in the test set. For these we find action-adverb pairs as follows.

We use the accompanying narrations to discover action-adverb pairs, for both training and testing. First we employ T-BRNN 
~\cite{tilk2016} to punctuate the subtitles\footnote{Note: YouTube ASR lacks punctuation}, then perform Part-of-Speech (POS) tagging with SpaCy's English core web model. 
We search for verb$\rightarrow$adverb relationships with the \textit{advmod} dependency, indicating the adverb modifies the verb. 
We exclude verbs with VBD (past tense) and VBZ (third person singular) tags 
as these correlate with actions not being shown in the video. 
For example, in `sprinkle some finely chopped coriander', `chopped' is tagged with VBD. 
Similarly, in `everything fits together neatly', the verb `fits' is tagged as VBZ. Examples of the (action, adverb) pairs obtained from the pipeline with the corresponding video snippets are shown in Fig.~\ref{fig:dataset}.
Additionally, we manually filter actions and adverbs that are not visual, \eg `recommend' and `normally', respectively.
We explored automatic approaches such as word concreteness scores~\cite{brysbaert2014concreteness}, but found these approaches to be unreliable. 
We also group verbs into clusters to avoid synonyms as in~\cite{damen2018EPICKITCHENS}, \ie we consider `put' and `place' as the same action.
From this process, we obtain 15,266 instances of action-adverb pairs.

However, these have a long tail of adverbs that are mentioned only a handful of times. 
We restrict our learning to 6 commonly used adverbs, that come in 3 pairs of antonyms: `partially'/`completely', `quickly'/`slowly' and `finely'/`coarsely'. These adverbs appear in 263 unique action-adverb pairs with 72 different actions. 
We show the distribution of adverbs per action in Fig.~\ref{fig:action_dist}.
While our training is noisy, \ie~actions might not appear in the video (refer to Fig.~\ref{fig:dataset} bottom), we clean the test set for accurate evaluation of the method.
We only consider test set videos where the action-adverb is present in the video and appears within the 20 seconds around the narration timestamp.
These correspond to 44\% of the original test set, which is comparable to the 50\% level of noise reported by the authors in~\cite{miech2019howto100m}.

This results in 5,475 action-adverb pairs in training and 349 in testing.  
We consider the mean timestamp between the verb and adverb narrations as the weak supervision for the action's location.
These action-adverb weak timestamp annotations and accompanying code are publicly available\footnote{\url{https://github.com/hazeld/action-modifiers}}.

\section{Experiments}
\label{sec:results}
We first describe the implementation details of our method, followed by the metrics we use for evaluation. We then present our results against those of baselines and evaluate the contribution of the different components.

\medskip
\noindent\textbf{Implementation Details.}
We sample all videos at 25fps and scale to a height of 256 pixels. We use I3D~\cite{carreira2017quo} with 16 frame segments, pre-trained on Kinetics~\cite{kay2017kinetics},
for both RGB and optical flow. We concatenate these to create $2048$D features, extracted once  
per second as in~\cite{zhukov2019cross}, for $T=20$ seconds around the narration timestamp. 

In all experiments, our embedding space $E$ is $300$D, the same as the GloVe word representation~\cite{pennington2014glove}. 
We initialize the action embeddings with the verb's GloVe vector, pre-trained on the Wikipedia and Gigaword corpora. The action modifiers $O_m$ are initialized with the identity matrix such that they have no effect at first. For our scaled dot-product attention, $Q$ is of size $75\!\times\!1$ and $K$ and $V$ are of size $75\!\times\!T$. We use $4$ attention heads in $f'(x,a)$.

All our models are trained with the Adam optimizer~\cite{kingma2014adam} for 1000 epochs with a batch size of 512 and a learning rate of $10^{-4}$. To aid disentangling the actions and adverbs, we first let the model learn only actions (optimized by $\mathcal{L}_{triplet}$) for 200 epochs before introducing the action modifiers. The weights of the action modifiers $W_m$ (Eq.~\ref{eq:actionmodifier}) are then learned at a slower rate of $10^{-5}$.

\medskip
\noindent\textbf{Evaluation Metric.}
We report mean Average Precision~(mAP) for video-to-adverb and adverb-to-video retrieval. For \textbf{video-to-adverb} given a video and the narrated action we rank the 6 adverbs' relevance. For \textbf{adverb-to-video} given an adverb query (\eg `slowly'), we rank videos by the distance of each video labelled with its associated action (\eg `put') to the text embedding of the verb-adverb (\eg `put slowly') and calculate mAP across the 6 adverbs.

We also report mAP where we restrict the retrieval to the adverb and its antonym, which we refer to as the \textbf{Antonym} setting.
This `Antonym' metric better represents the given labels, therefore we use it for the ablation study. 
To clarify, we may have a video narrated `cut coarsely'. We are thus confident the cut was not performed `finely', however we cannot judge the speed of (`quickly' or `slowly').
In Antonym video-to-adverb, there are only two possible adverbs to retrieve, thus we report Precision@1 (P@1) which is the same as binary classification accuracy.
Similarly, we report mAP Antonym for adverb-to-video retrieval, where we only rank videos labeled with the adverb or its antonym.



\subsection{Comparative Results}
\label{sec:compres}
We first compare our work to baselines.
Since ours is the first work to learn from adverbs in videos, we adapt methods that learn attributes of objects in images for comparison, as this is the most similar existing task to ours. In this adaptation, actions replace objects, and adverbs replace attributes/adjectives.

We compare to RedWine~\cite{misra2017red} and AttributeOp~\cite{nagarajan2018attributes} as well as the LabelEmbed baseline proposed in~\cite{misra2017red} which uses GloVe features in place of SVM classifier weights.
We replace the image representation by a uniformly weighted visual representation of video segments.
Similar to our evaluation, we report results when the action is given in testing, referred to as the `oracle' evaluation in~\cite{nagarajan2018attributes}. Furthermore, for a fair comparison, we use only the antonym as the negative in each method's loss, as we do in Eq.~\ref{eq:lossadverb}.
AttributeOp proposes several linguistic inspired regularizers; we report the best combination of regularizers for our dataset --- the auxiliary and commutative regularizers. We also compare to random chance and a naive binary classifier per adverb pair. This classifier is analogous to the Visual Product baseline used in~\cite{misra2017red, nagarajan2018attributes}. 
We report on both versions of this baseline, a Linear SVM which trains a binary one-vs-all classifier per adverb (Classifier-SVM) and a 6-way MLP of two fully connected layers (Classifier-MLP). In video-to-adverb, we rank adverbs by classifiers' confidence scores, as in~\cite{nagarajan2018attributes}. In adverb-to-video, we use the confidence of the corresponding classifier or MLP output to rank videos.

\begin{table}[t]
\begin{center}
{\def\arraystretch{1.3}\tabcolsep=6.2pt
\begin{tabular}{@{}lcccc@{}}
\toprule
\multirow{2}{*}{Method} & \multicolumn{2}{c}{video-to-adverb} & \multicolumn{2}{c}{adverb-to-video} \\
\cmidrule(l){2-3} \cmidrule(l){4-5}

 & Antonym & All & Antonym & All \\
\midrule
Chance & 0.500 & 0.408 & 0.511 & 0.170\\ 
Classifier-SVM & 0.605 & 0.532 & 0.563 & 0.264\\
Classifier-MLP & 0.685 & 0.602 & 0.603 & 0.304\\
\hline
RedWine~\cite{misra2017red} & 0.693 & 0.594 & 0.595 & 0.290\\
LabelEmbed~\cite{misra2017red} & 0.717 & \underline{0.621} & \underline{0.618} & 0.297\\
AttributeOp~\cite{nagarajan2018attributes} & \underline{0.728} & 0.612 & 0.597 & \textbf{0.350}\\
\midrule
Ours & \textbf{0.808} & \textbf{0.719} & \textbf{0.657} & \underline{0.329}\\
\bottomrule
\end{tabular}
}
\end{center}
\caption{Comparative Evaluation. Best performance in \textbf{bold} and second best \underline{underlined}. We report results for both video-to-adverb and adverb-to-video retrieval with results restricted to the adverb and its antonym (Antonym) and when unrestricted (All).
}
\vspace*{-12pt}
\label{tab:t}
\end{table}

Comparative results are presented in Table~\ref{tab:t}. Our method outperforms all baselines for video-to-adverb retrieval, both when comparing against all adverbs and when restricting the evaluation to antonym pairs. 
We see that AttributeOp is the best baseline method, generally performing better than both RedWine and LabelEmbed. The two latter methods work on a fixed visual feature space, thus are prone to errors when the features are non-separable in that space. We can also see that LabelEmbed performs better than RedWine across all metrics, demonstrating that GloVe features are better representations than SVM classifier weights.
While AttributeOp marginally outperforms our approach on adverb-to-video `All', it underperforms on all other metrics, including our main objective, estimating the correct adverb over its antonym for a video query.

\label{sec:qual}
\begin{figure*}
    \centering
    \includegraphics[width=\linewidth]{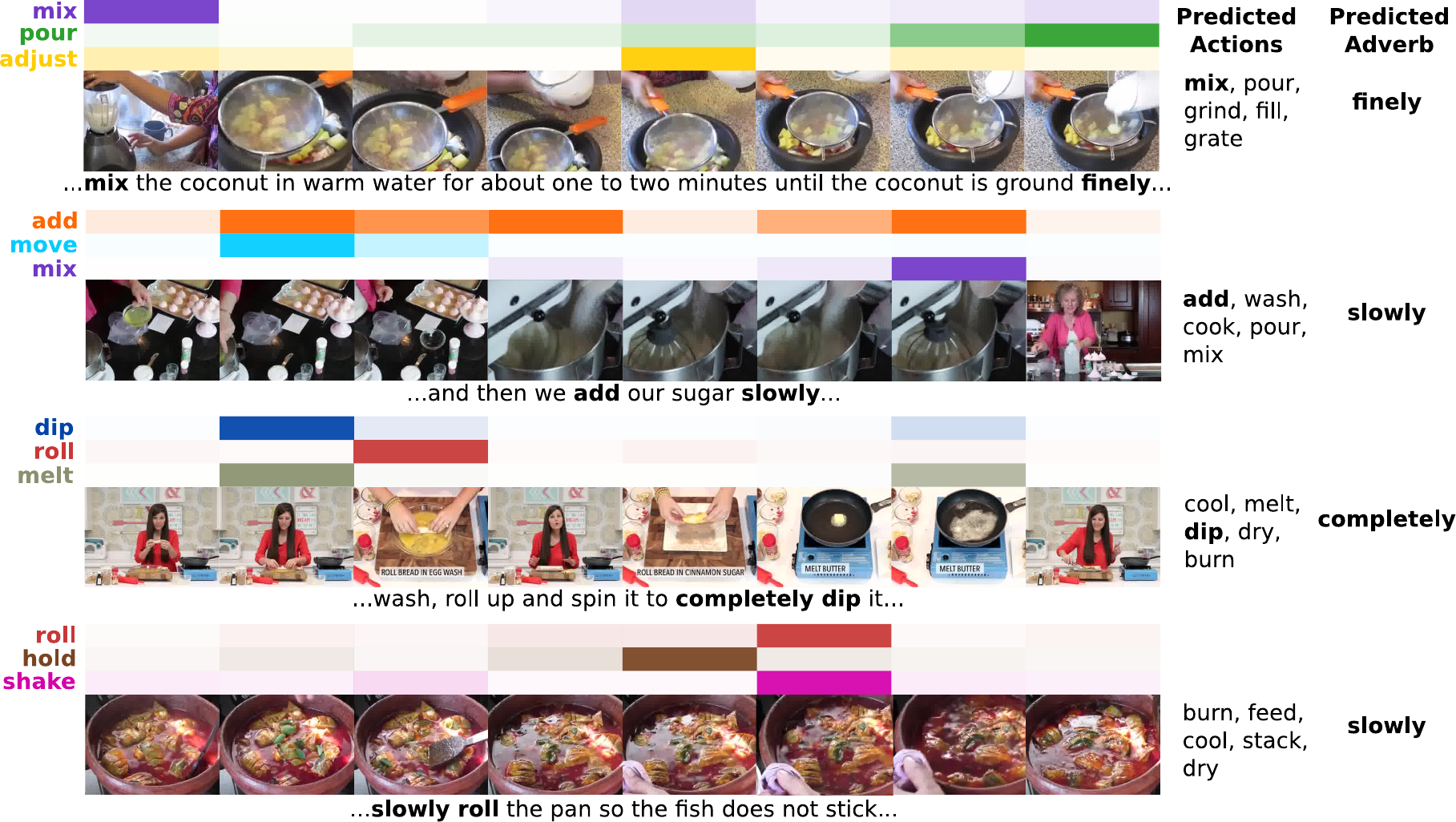}
    \caption{\textbf{Qualitative Results.} Temporal attention values from several action queries. The intensity of the color indicates the attention value. Recall that we use the narrated action to weight the relevance of video segments. Using that, we display the top-5 predicted actions, as well as the correctly predicted adverb for all cases.
    }
    \vspace*{-10pt}
    \label{fig:attention}
\end{figure*}

\subsection{Qualitative Results}


Fig.~\ref{fig:attention} presents video examples. For each, we demonstrate attention weights for several action queries. Our method is able to successfully attend to segments relevant to various query actions.
The figure also shows predicted actions, and predicted adverb when using the ground-truth action as the query. Our method is able to predict the correct adverb. In the last example, predicted actions are incorrect, but the method correctly identifies a relevant segment and that the action was done `slowly'.
We provide further insights into the learned embedding space in supplementary.

\subsection{Ablation Study}
We report 4 ablation studies on the various aspects of the method: the choice of action modifier transformation $O_m(\cdot)$, our scaled dot-product attention, the contributions of the loss functions, and the length of the video ($T$). 
We focus on video-to-adverb retrieval in the ablation using the Antonym P@1 metric, as this allows us to answer questions like: ``was the `cut' performed `quickly' or `slowly'?''.
\begin{table}[t]
\begin{center}
{\def\arraystretch{1.3}\tabcolsep=3pt
\begin{tabular}{@{}lccc@{}}
\toprule
$O_m(z)=$ & Dimension &Learned &P@1 \\ 
\midrule
$z + \mathrm{GloVe}(m)$ &1D & & 0.735\\
$z + b_m$ &1D &$\checkmark$ &0.749\\
\midrule
$W_m z$ &2D &$\checkmark$ & \textbf{0.808}\\
$W_{m_2}\mathrm{ReLU}(W_{m_1} z + b_{m})$ &2D &$\checkmark$ & 0.742\\
\bottomrule
\end{tabular}
}
\end{center}
\vspace{-0.3em}
\caption{Comparison of action modifier representation $O_m(\cdot)$. 
The linear transformation choice clearly improves results.}
\vspace{-12pt}
\label{tab:adverb_rep}
\end{table}

\medskip
\noindent\textbf{Action Modifier Representation.}
In Table~\ref{tab:adverb_rep} we examine different representations for the action modifiers $O_m(\cdot)$ (Eq.~\ref{eq:actionmodifier}). 
We compare to a fixed translation from the GloVe representation of the adverb ($m$), which is not learned, to three learned representations.
First, a learned translation vector $b_m$ 
initialized from the GloVe embedding is used. 
Second, our chosen representation - a 2D linear transformation with matrix $W_m$ as in Eq.~\ref{eq:actionmodifier}.
Third, we learn a non-linear transformation implemented as two fully connected layers, the first with a ReLU activation. 

Results show the linear transformation clearly outperforms a vector translation or the non-linear transformation. The translation vector does not having enough capacity to represent the complexity of the adverb, while the nonlinear transform is prone to over-fitting.



\medskip
\noindent\textbf{Temporal Attention.} 
In Table~\ref{tab:temporal_agg}, we compare our proposed multi-head scaled dot-product attention (Sec.~\ref{sec:webly}) with alternative approaches to temporal aggregation and attention. In this comparison, we also report action retrieval results, with video-to-action mAP. That is, given the embedding of the video $f'(x,a)$ queried by the ground-truth action, we rank all actions in the embedding $\forall a: g(a)$ by their distances to the visual query and evaluate the rank of the correct action. Our method does not aim for action retrieval as it assumes knowledge of the ground-truth action, however this metric evaluates the quality of the weakly supervised embedding space. Results are compared to: 

\vspace*{-6pt}
\begin{itemize}[itemsep=-0.3ex]
\item \textbf{Single}: uses only a one-second clip at the timestamp.
\item \textbf{Average}: uniformly weights the $T$ features.
\item \textbf{Attention from~\cite{long2018attention}}:  widely used class agnostic attention, calculating attention with two fully connected layers, $f'(x,a) = \sigma(w_1 \tanh(W_2f(x)))W_3 f(x).  $  
\item \textbf{Class-specific Attention}: a version of the above with one attention filter per action class.
\item \textbf{Ours w/o two-stage optimization}: our attention without the first 200-epoch stage of learning action triplets without learning adverbs/modifiers. 
\item \textbf{Ours}: our attention as described in Sec.~\ref{sec:webly}.
\end{itemize}

\vspace*{-6pt}

\noindent Table~\ref{tab:temporal_agg} demonstrates superior performance of our method for the learning of action embeddings and, as a consequence, better learning of action modifiers. These results also demonstrate the challenge of weak-supervision, with video-to-action only performing at 0.246 mAP when considering only one second surrounding the narrated action. This improves to 0.692 with our method.


\begin{table}[t]
\begin{center}
{\def\arraystretch{1.2}\tabcolsep=10.2pt
\begin{tabular}{@{}lcc@{}}
\toprule
 Method & Action &  Adverb \\ 
\midrule
Single & 0.246 & 0.705\\
Average & 0.257 & 0.716\\
Attention from~\cite{long2018attention}  & 0.235 & 0.708 \\
Class-specific Attention& 0.401 & 0.728\\
 \midrule
Ours w/o two-stage optimization & 0.586 & 0.774\\
Ours & {\bfseries 0.692} & {\bfseries 0.808} \\
\bottomrule
\end{tabular}
}
\end{center}
\vspace{-0.2em}
\caption{Comparison of temporal attention methods. We report video-to-action retrieval mAP and video-to-adverb retrieval P@1.
}
\label{tab:temporal_agg}
\end{table}
\medskip
\noindent\textbf{Loss Functions.} 
We also evaluate the need for two separate loss functions (Eqs.~\ref{eq:lossaction} and~\ref{eq:lossadverb}).
As an alternative approach we use a single loss where the negative contains a different action, a different adverb or both. 
This performs worse by 0.03 P@1. 
Using both losses, but with another adverb as opposed to only the antonym $\overline{m}$ in Equation~\ref{eq:lossadverb} also results in worse performance by 0.04 P@1. 
\begin{figure}
    \centering
    \includegraphics[width=0.90\linewidth]{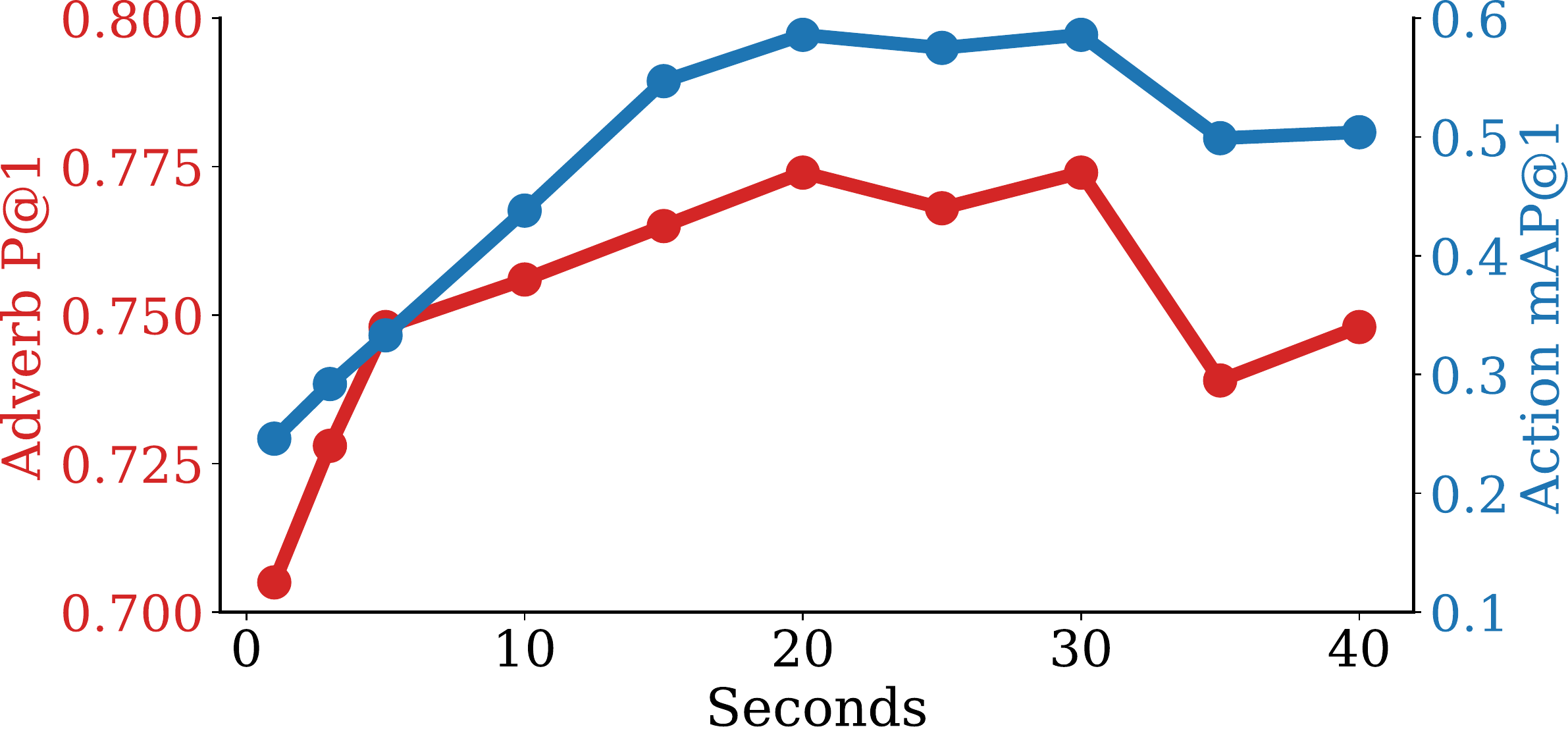}
    \caption{Performance as $T$ increases. Blue (axis and plot) shows video-to-action retrieval mAP while red shows video-to-adverb retrieval with Antonym P@1. 
    }
    \label{fig:t}
    \vspace{-8pt}
\end{figure}

\medskip
\noindent\textbf{Effect of \boldmath{$T$}}.
In Fig.~\ref{fig:t}, we evaluate how the length of the video ($T$) extracted around the weak timestamp affects the model (Sec.~\ref{sec:webly}). For larger $T$, videos are more likely to contain the relevant action, but also other actions. Our embedding function $f'(x,a)$ is able to ignore other actions in the video, up to a point, and successfully learn to attend to the relevant parts given the query action, resulting in better performance with $T\in \{20\mathrel{{.}\,{.}}\nobreak 30\}$. 


\medskip
\noindent \textbf{Comparison with Action Localization}.
In this work, we perform weakly supervised embedding to learn action modifiers by attending to action relevant segments. Here, we test whether weakly supervised action localization can be used instead of our proposed attention, to locate key segments before learning action modifiers.

We use published code of two state-of-the-art weakly supervised action localization methods: W-TALC~\cite{paul2018w} and CMCS~\cite{liu2019completeness}. First, we test the output of these methods with a binary adverb-antonym classifier (Classifier-MLP as in Sec.~\ref{sec:compres}).
We also test these methods in combination with our embedding and action modifier transformations. 
For this, we use the methods' predicted action-relevant segments, and average their representation to replace $f'(x,a)$~(Avg). Finally, we combine these relevant segments with our scaled dot-product attention~(SDP).

From Table~\ref{tab:act_loc} we can conclude that using the output of a weakly-supervised localization method is insufficient, and our joint optimization performs best. Worth noting, localizing the action using W-TALC followed by averaging relevant segments outperforms averaging all segments (0.739 vs. 0.716 from Table~\ref{tab:temporal_agg}). This shows that W-TALC is capable of finding some relevant segments. This is further improved by our scaled dot-product attention.

\begin{table}
\begin{center}
{\def\arraystretch{1.3}\tabcolsep=6.2pt
\begin{tabular}{@{}lcll@{}}
\toprule
Method & Attention & Adverb Rep & P@1 \\ 
\midrule
\multirow{3}{*}{W-TALC~\cite{paul2018w}} &Avg  & Classifier-MLP & 0.705\\
&Avg  & Action Modifiers & 0.739\\
&SDP & Action Modifiers & 0.768 \\
\midrule
\multirow{3}{*}{CMCS~\cite{liu2019completeness}} &Avg & Classifier-MLP  & 0.696\\
&Avg  & Action Modifiers & 0.699\\
&SDP & Action Modifiers & 0.705\\ 
\midrule
Ours  & SDP & Action Modifiers & \textbf{0.808}\\
\bottomrule
\end{tabular}
}
\end{center}
\vspace{-0.2em}
\caption{Comparison of our method (Ours) to weakly supervised action localization methods, with and without our scaled dot-product (SDP) and action modifier representations. 
} 
\vspace{-12pt}
\label{tab:act_loc}
\end{table}


\section{Conclusion}
\label{sec:conc}
This paper presents a weakly supervised method to learn from adverbs in instructional videos. Our method learns to obtain and embed the relevant part of the video with scaled dot product attention, using the narrated action as a query. The method then learns action modifiers as linear transformations 
on the embedded actions; shared between actions. 
We train and evaluate our method on parsed action-adverb pairs sourced from YouTube videos of 83 tasks.
Results demonstrate that our method outperforms all baselines, achieving 0.808 mAP for video-to-adverb retrieval, when considering the adverb versus its antonym. 

Future work will involve learning from few shot examples in order to represent a greater variety of adverbs as well as exploring applications to give feedback to people guided by instructional videos or written instructions. 

\medskip
\noindent\textbf{Acknowledgements:} Work is supported by an EPSRC DTP, EPSRC GLANCE (EP/N013964/1), Louis Vuitton ENS Chair on Artificial Intelligence, the MSR-Inria joint lab and the French government program, reference ANR-19-P3IA-0001 (PRAIRIE 3IA Institute). Part of this work was conducted during H.~Doughty's internship at INRIA Willow Team. Work uses publicly available dataset.

{\small
\bibliographystyle{ieee_fullname}
\bibliography{egbib}
}
\clearpage
\newpage

\appendix
\section*{Overview of Supplementary Material}
This supplementary document allows us to examine a sample of the embedding space (Appendix~\ref{app:emb}).
We present two additional ablations. First, per adverb results across modalities, giving additional insight into the learning of adverbs (Appendix~\ref{app:per}). We also investigate alternative choices for the query $Q$ to the scaled dot-product attention (Appendix~\ref{app:q}). 

\section{Embedding Space}
\label{app:emb}
\begin{figure*}[t]
   \centering
   \includegraphics[width=0.75\linewidth]{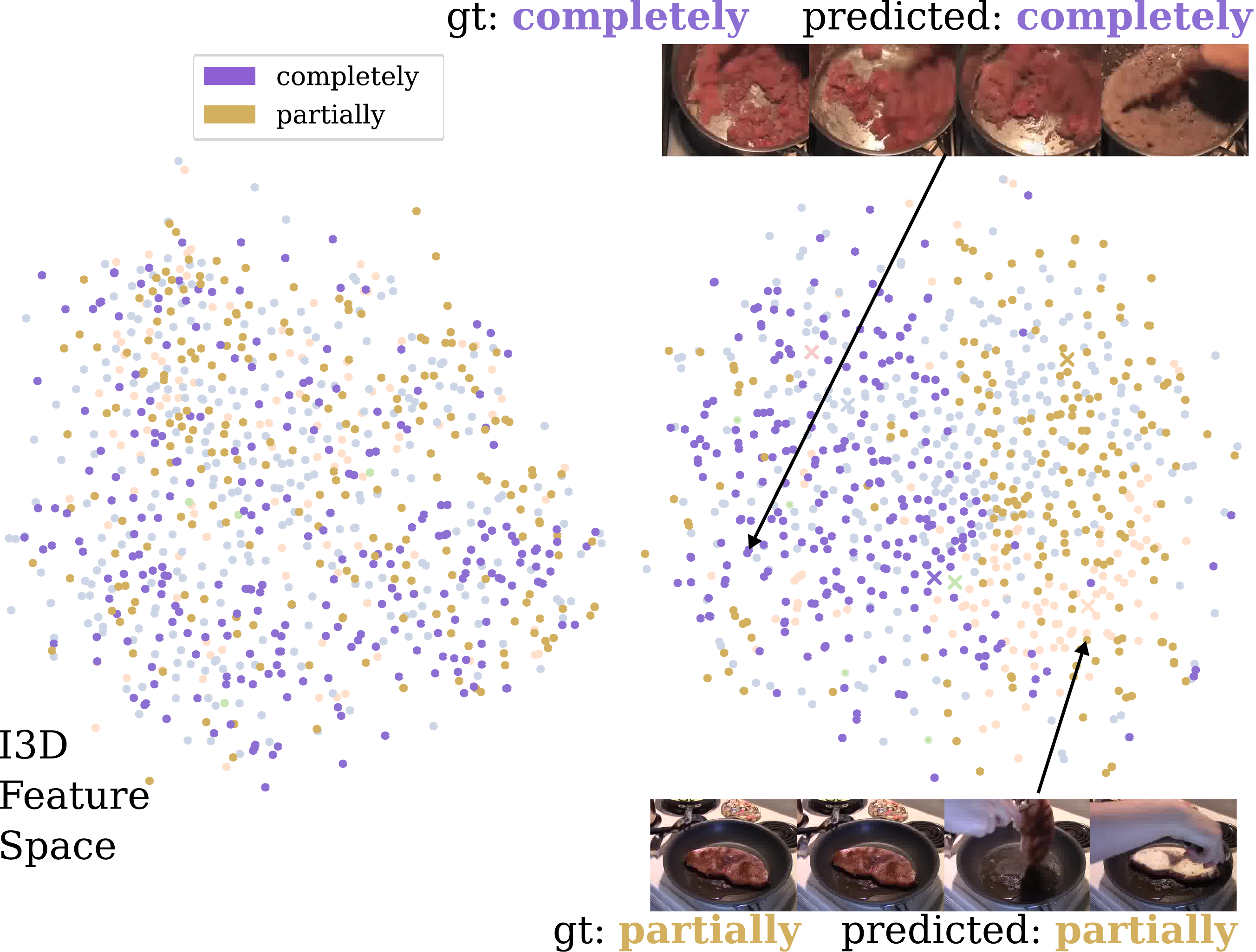}
   \caption{Comparison between the feature space for the action `cook' before and after training highlighting antonym pairs. We highlight the `completely'/`partially' pair with the other adverbs faded out. 
   }
   \label{fig:embedding_partially}
\end{figure*}

\begin{figure*}[t]
   \centering
   \includegraphics[width=0.75\linewidth]{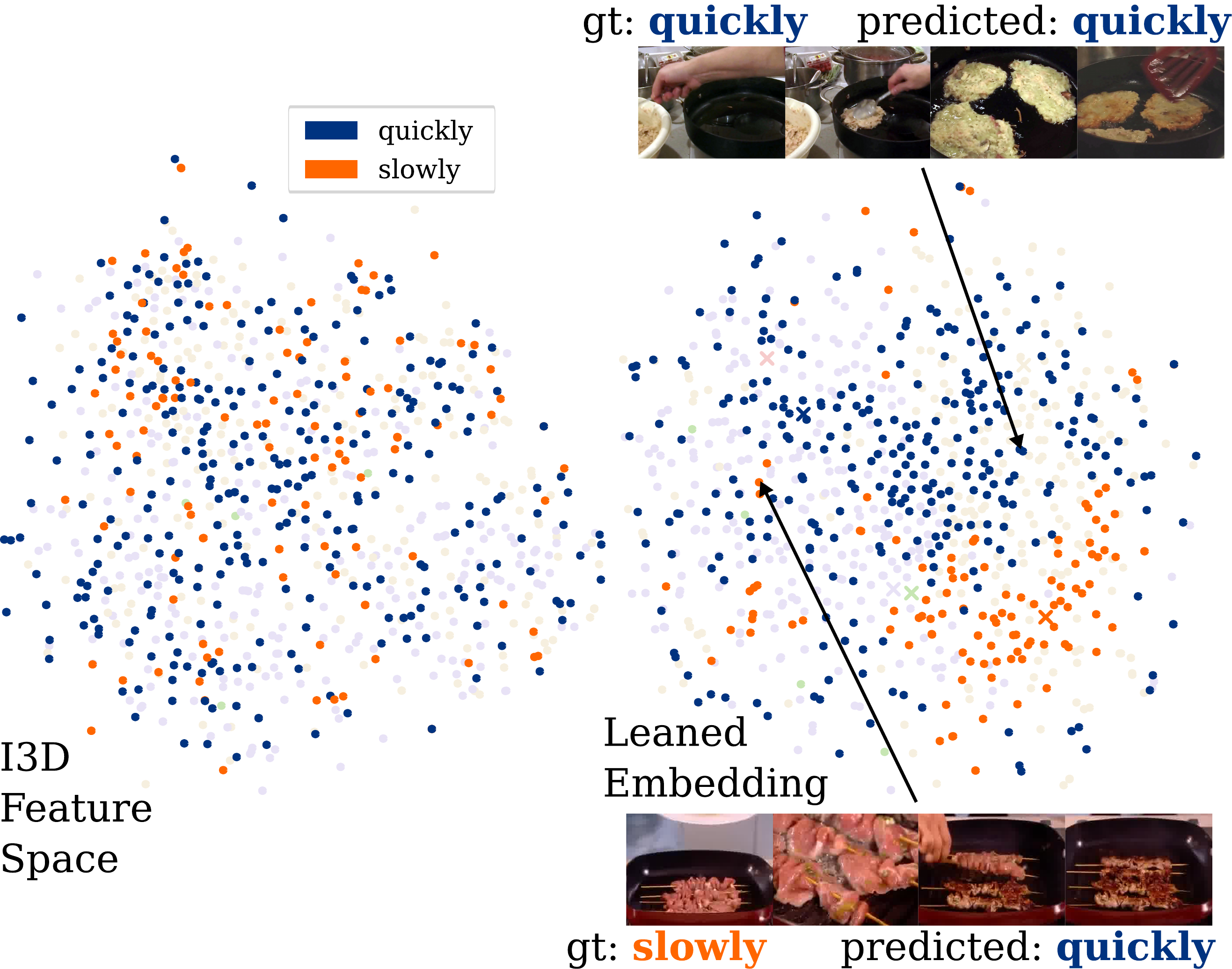}
   \caption{Comparison of the features spaces before and after training for the antonym pair `quickly'/`slowly' in the action `cook'. We fade out adverbs which are not `quickly' or `slowly'.
   }
   \label{fig:embedding_quickly}
\end{figure*}

While visualizing the high-dimensional embedding space is difficult, we provide t-SNE projections of this space for a sample, to show the learning achieved.
We consider all videos of the narrated action `cook', and show the embedding space before (\ie from I3D features) and after training. 
We highlight in two figures
adverb-antonym pairs `completely'/`partially' (Fig.~\ref{fig:embedding_partially}) and `quickly'/`slowly' (Fig.~\ref{fig:embedding_quickly}) and fade out points corresponding to other adverbs for ease of viewing. 
In each case, we show that our training successfully separates the embedding space based on the adverb.
From this figure we can see that the method successfully separates adverbs from their antonyms. 
The figure also visualize a couple of video examples in each case, with 3 videos correctly embedded with the corresponding ground-truth and one incorrect prediction `slowly'$\rightarrow$`quickly'.

\begin{figure*}[t]
   \centering
   \includegraphics[width=0.9\linewidth]{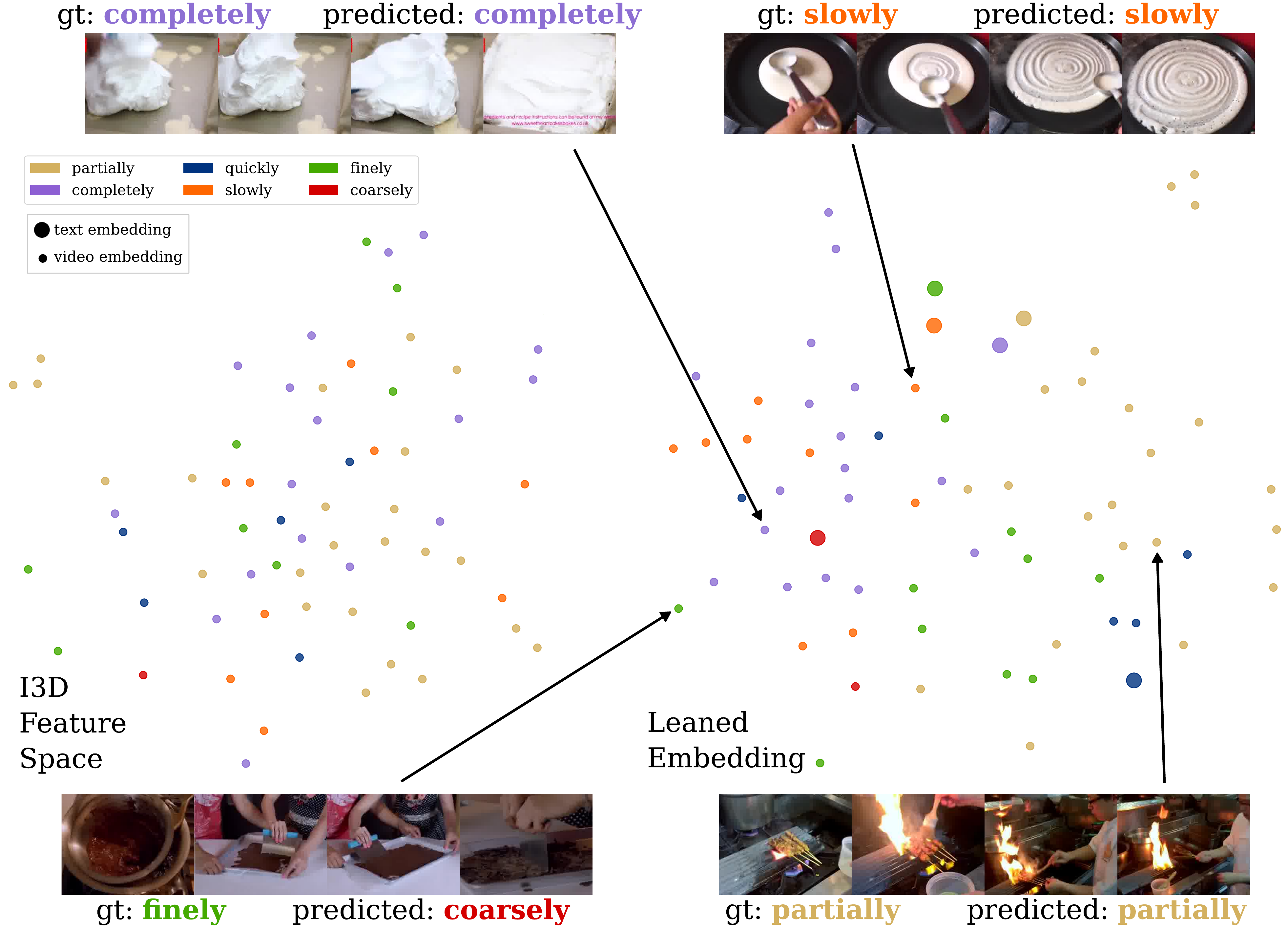}
   \caption{Comparison of the features spaces before and after training for the action `spread'. All adverbs are shown.
   }
   \vspace{26in}
   \label{fig:spread}
\end{figure*}

Similarly, we plot the t-SNE projections of the embedding for videos narrated with the action `spread' (Fig.~\ref{fig:spread}) before and after training. From this we can see that despite having much fewer examples of the action, the method is still able to successfully separate adverb-antonym pairs.

\section{Per Adverb Results}
\label{app:per}
\begin{figure}
   \centering
   \includegraphics[width=\linewidth]{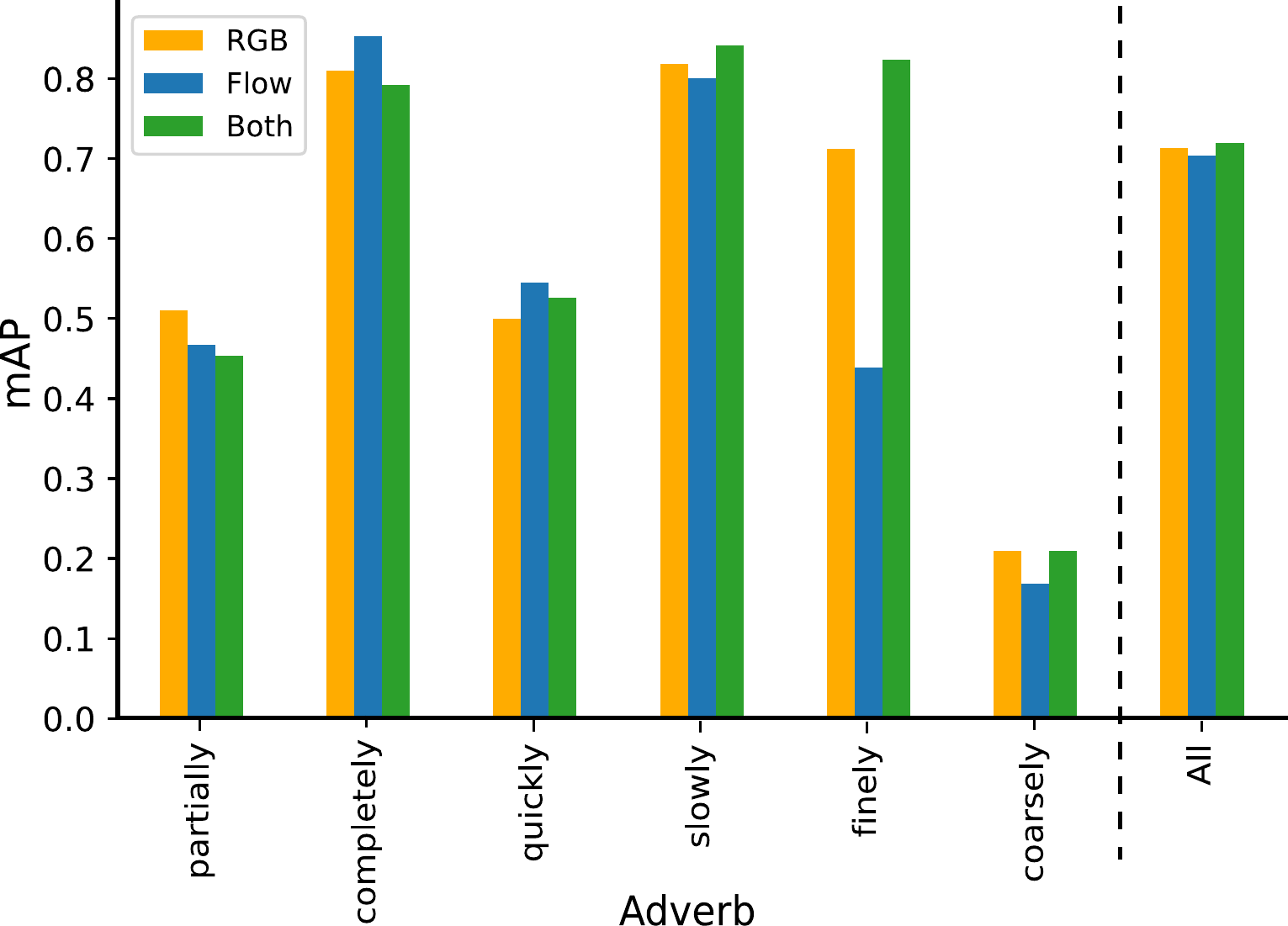}
   \caption{Video-to-adverb retrieval mAP per adverb with different modalities.}
   \label{fig:per_adverb}
\end{figure}

In Figure~\ref{fig:per_adverb} we show the effect of different modalities on the results per adverb.
Firstly, we observe that considering all adverbs (All), the inclusion of both RGB and Flow is better than either modality separately. However, modalities perform differently across individual adverbs. For example, `finely' is retrieved significantly more successfully with RGB than with Flow. Unsurprisingly, `quickly' and `slowly' benefit from the inclusion of Flow features alongside RGB.


\section{Choice of $Q$}
\label{app:q}
As noted in the main manuscript, we use a query $Q$ to attend to the relevant parts of the video, for weakly-supervised embedding.
We have chosen the embedding of the action, $g(a)$, as the query to our scaled dot-product attention (Eq.~\ref{eq:att}). Our attention is calculated by the compatibility of the query $Q$ with the key $K$ (a linear projection of the video segment features), therefore the choice of $Q$ is integral to the weakly-supervised embedding. 
Here, we compare $Q = W^Q g(a)$ to several alternatives, including incorporating the adverb into the query.
We report the results in Table~\ref{tab:q}.
For this ablation, we do not use the two-stage optimization, and thus the performance matches that of 0.774 in Table 3 in the paper.

First, we compare the action's embedding $g(a)$ to a one-hot vector of the action. The embedding offers a better query.
Second, we test using the adverb as a query. In this case, we use a single adverb from each antonym pair (\eg `slowly'/`quickly'$\rightarrow$`quickly'). This offers an understanding of the type of adverb we are after, so as to pick relevant video segments to this action manner. We compare the GloVe representation to a flattened representation of the learned action modifier. Again, while this 
allows the method to focus on segments relevant to the type of action manner,
using the embedding of the action performs best.
Finally, we test the full action-adverb embedding $O_m(g(a))$.
This showed a drop in performance compared to using the action's embedding alone. This is potentially related to the fact that adverbs are not mutually exclusive as described in the paper's results.

\begin{table}
\begin{center}
{\def\arraystretch{1.3}\tabcolsep=10.2pt
\begin{tabular}{@{}llll@{}}
\toprule
& $Q$ & P@1\\ 
\midrule
\multirow{2}{*}{Action} & $g(a)$ & \textbf{0.774} \\
& One-hot Vector & 0.736\\
 &0.456\\
\midrule
\multirow{2}{*}{Adverb} & GloVe & 0.702 \\
& $\text{Vec}(W_m)$ & 0.731 \\
\midrule
\multirow{1}{*}{Both} 
& $O_m(g(a))$ & 0.728\\
\bottomrule
\end{tabular}
}
\end{center}
\vspace{-0.2em}
\caption{Comparison of the choice of $Q$.} 
\label{tab:q}
\vspace{-0.6em}
\end{table}

\end{document}